\documentclass[final]{elsarticle}
\pdfoutput=1
\usepackage[export]{adjustbox}

\usepackage{hyperref}
\usepackage{lineno}
\usepackage{times}
\usepackage{amssymb}
\usepackage{latexsym}
\usepackage{colortbl}
\usepackage[textsize=small]{todonotes}

\usepackage[inline,shortlabels]{enumitem}
\usepackage{xcolor}
\usepackage{amsmath}
\usepackage{booktabs}
\usepackage{array,multirow}
\usepackage{caption}
\usepackage{comment}
\usepackage{pifont}
\usepackage{tikz}
\usepackage{siunitx}
\usepackage{subcaption}

\graphicspath{{figures/}}

\makeatletter
\let\UrlSpecialsOld\UrlSpecials
\def\UrlSpecials{\UrlSpecialsOld\do\/{\Url@slash}\do\_{\Url@underscore}}%
\def\Url@slash{\@ifnextchar/{\kern-.11em\mathchar47\kern-.2em}%
    {\kern-.0em\mathchar47\kern-.08em\penalty\UrlBigBreakPenalty}}
\makeatother

\usepackage{color}

\newcommand{\yesmarker}{\cmark}
\newcommand{\nomarker}{$-$}

\newcommand{\eg}{e.g., }
\newcommand{\ie}{i.e., }
\newcommand{\vs}{vs.\ }
\newcommand{\figref}[1]{Fig.~\ref{#1}}    
\newcommand{\Figref}[1]{Figure~\ref{#1}}  
\newcommand{\tabref}[1]{Table~\ref{#1}}
\newcommand{\Tabref}[1]{Table~\ref{#1}}
\newcommand{\secref}[1]{Section~\ref{#1}}
\newcommand{\eqs}{Eqs.}
\newcommand{\eq}{Eq.}
\newcommand{\equref}[1]{\eq~\ref{#1}}
\newcommand{\equsrefrange}[2]{\eqs~\ref{#1}--\ref{#2}}

\newcommand{\cmark}{\ding{51}}
\newcommand{\xmark}{\ding{55}}

\newcommand\LSTM{\mbox{LSTM}} 
\newcommand\LSTMs{{\LSTM}s} 
\newcommand\biLSTM{Bi\LSTM}
\newcommand\biLSTMs{Bi{\LSTM}s} 
\newcommand\TreeLSTM{Tree-\LSTM}
\newcommand\TreeLSTMs{Tree-{\LSTM}s} 
\newcommand\BLSTM{B-\LSTM}
\newcommand\TLSTM{T-\LSTM}
\newcommand\NTLSTM{NT-\LSTM}

\newcommand\MaxILP{ILP Coverage}
\newcommand\TopILP{ILP Naive}

\definecolor{wordcolor}{HTML}{36a9e0}
\definecolor{embeddingcolor}{HTML}{f9b233}
\definecolor{biLSTMcolor}{HTML}{e84e1b}
\definecolor{attentioncolor}{HTML}{a2195b}
\definecolor{hiddencolor}{HTML}{008d36}

\definecolor{darkspringgreen}{rgb}{0.09, 0.45, 0.27}
\colorlet{positive0001}{darkspringgreen!75}
\colorlet{positive001}{darkspringgreen!50}
\colorlet{positive01}{darkspringgreen!25}

\definecolor{deepcarmine}{rgb}{0.66, 0.13, 0.24}

\colorlet{negative0001}{deepcarmine!75}
\colorlet{negative001}{deepcarmine!50}
\colorlet{negative01}{deepcarmine!25}

\biboptions{authoryear}

\journal{Expert Systems with Applications}

\usepackage{etoolbox}
\patchcmd{\pprintMaketitle}
 {\ifvoid\absbox\else\unvbox\absbox\par\vskip10pt\fi}
 {\ifvoid\absbox\else\clearpage\unvbox\absbox\par\vskip30pt\fi}
 {}{}
\patchcmd{\pprintMaketitle}
 {\hrule\vskip12pt}
 {}
 {}{}
\patchcmd{\pprintMaketitle}
 {\hrule\vskip12pt}
 {}
 {}{}
\appto{\pprintMaketitle}{\clearpage}

\begin{document}

\begin{frontmatter}
\title{Solving Arithmetic Word Problems by Scoring Equations with Recursive Neural Networks}

\author[ugent]{Klim~Zaporojets\corref{cor1}}
\ead{klim.zaporojets@ugent.be}
\author[vub]{Giannis~Bekoulis\corref{cor2}}
\ead{gbekouli@etrovub.be}
\author[ugent]{Johannes~Deleu}
\ead{johannes.deleu@ugent.be}
\author[ugent]{Thomas~Demeester}
\ead{ thomas.demeester@ugent.be}
\author[ugent]{Chris~Develder}
\ead{chris.develder@ugent.be}
\cortext[cor1]{Corresponding author}
\cortext[cor2]{The work presented in the paper was performed while dr.\ Bekoulis was with Ghent University -- imec,  IDLab, Department of Information Technology.}

\address[ugent]{Ghent University -- imec, IDLab, Dept.\ of Information Technology (INTEC),\\
Technologiepark Zwijnaarde 15, 9052 Ghent, Belgium}

\address[vub]{Vrije Universiteit Brussel -- imec, Dept.\ of Electronics and Informatics (ETRO),\\
Pleinlaan 9, 1050 Brussels, Belgium}

\begin{abstract}
  Solving arithmetic word problems is a cornerstone task in assessing language understanding and reasoning capabilities in NLP systems. Recent works 
  use automatic extraction and ranking of candidate solution equations providing the answer to arithmetic word problems. In this work, we explore novel approaches to score such candidate solution equations using tree-structured
  recursive neural network (Tree-RNN) configurations. 
  The advantage of this Tree-RNN approach over using more established sequential representations, is that it can naturally capture the structure of the equations. Our proposed method consists of transforming the mathematical expression of the equation into an expression tree. Further, we encode this tree into a Tree-RNN by using different \TreeLSTM{} architectures.  
   Experimental results show that our proposed method 
   \begin{enumerate*}[(i)]
    \item  improves overall performance with more than 3\% accuracy points compared to previous state-of-the-art, and with over 15\% points on a subset of problems that require more complex reasoning, and \item outperforms sequential LSTMs by 4\% accuracy points on such more complex problems. 
\end{enumerate*}
\end{abstract}

\end{frontmatter}

\section{Introduction}
\label{sec:intro}

\begin{figure}[b]
\centering
\includegraphics[width=0.7\columnwidth]{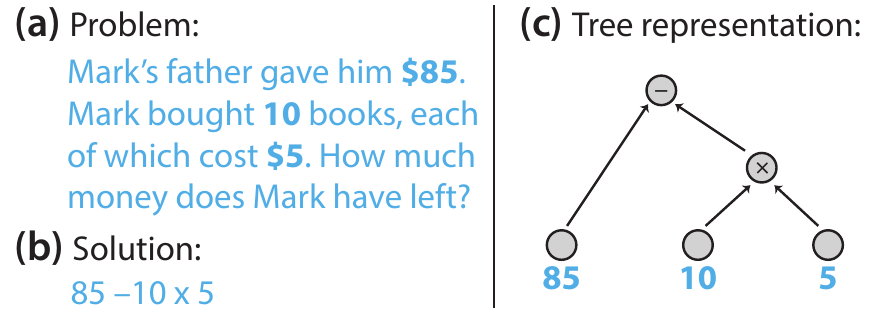}
\captionsetup{singlelinecheck=off}
\caption[test]{
An example of arithmetic word problem from the SingleEQ dataset. It illustrates the 
\protect\begin{enumerate*}[(a)]
    \item\label{fig1a} \textit{an arithmetic word problem} statement,
    \item\label{fig1b} the respective \textit{solution} formula, and
    \item\label{fig1c} the \textit{expression tree} representing the solution.
\protect\end{enumerate*}
}
\label{fig:example}
\end{figure}

\noindent Natural language understanding often requires the ability to comprehend and reason with expressions involving numbers. This has produced a recent rise in interest to build applications to automatically solve math word problems~\citep{kushman2014learning,koncel2015parsing,mitra2016learning,wang2018mathdqn,zhang2019gap}. These math problems consist of a textual description comprising numbers with a question that will guide the reasoning process to get the numerical solution (see \figref{fig:example} for an example). 
This is a complex task because of
\begin{enumerate*}[label=(\roman*)]
    \item \label{problem1} the large output space of the possible equations representing a given math problem, and
    \item \label{problem2} reasoning required to understand the problem.
\end{enumerate*} 

The research community has focused in solving mainly two types of mathematical word problems: \textit{arithmetic word problems} 
\citep{hosseini2014learning, mitra2016learning, wang2017deep, li2019modeling, chiang2019semantically} 
and \textit{algebraic word problems} \citep{kushman2014learning, shi2015automatically, ling2017program, amini2019mathqa}. Arithmetic word problems 
can be solved using basic mathematical operations ($+, -, \times, \div$) and 
involve a single unknown variable. Algebraic word problems, on the other hand, 
involve more complex operators such as square root, exponential and logarithm with multiple unknown variables. 
In this work, we focus on solving \textit{arithmetic word problems} such as the one illustrated in \figref{fig:example}. This figure illustrates \begin{enumerate*}[label=(\alph*)]
    \item \textit{arithmetic word problem} statement, 
    \item the arithmetical formula of the \textit{solution} to the problem, and 
    \item the \textit{expression tree} representation of the solution formula where the leaves are connected to quantities and internal nodes represent operations
\end{enumerate*}. 

The main idea of this paper is to explore the use of tree-based Recursive Neural Networks (Tree-RNNs) to encode and score the expression tree (illustrated in \figref{fig:example}\ref{fig1c} that represents a candidate arithmetic expression of a specific arithmetic word problem). 
This contrasts with predominantly sequential neural representations \citep{wang2017deep, wang2018translating, chiang2019semantically} that encode the problem statement from left to right or vice versa.
By using Tree-RNN architectures, we can naturally embed the equation inside a tree structure such that the link structure directly reflects the various mathematical operations between operands selected from the sequential textual input. We hypothesize that this structured approach can efficiently capture the semantic representations of the candidate equations to solve more complex arithmetic problems involving multiple and/or non-commutative operators. To test our results, we use the recently introduced SingleEQ dataset \citep{koncel2015parsing}. It contains a collection of 508 arithmetic word problems with varying degrees of complexity. This allows us to track the performance of the evaluated systems on subsets that require different reasoning capabilities. More concretely, we subdivide the initial dataset into different subsets of varying reasoning complexity (\ie based on the number of operators, commutative (symmetric) or non-commutative (asymmetric) operations), to investigate whether  
the performance of the proposed architecture remains consistent across problems of increasing complexity.

\begin{figure}[!ht]
\centering
\includegraphics[width=1.0\columnwidth,trim={0.5cm 7.0cm 7.7cm 0.2cm},clip]{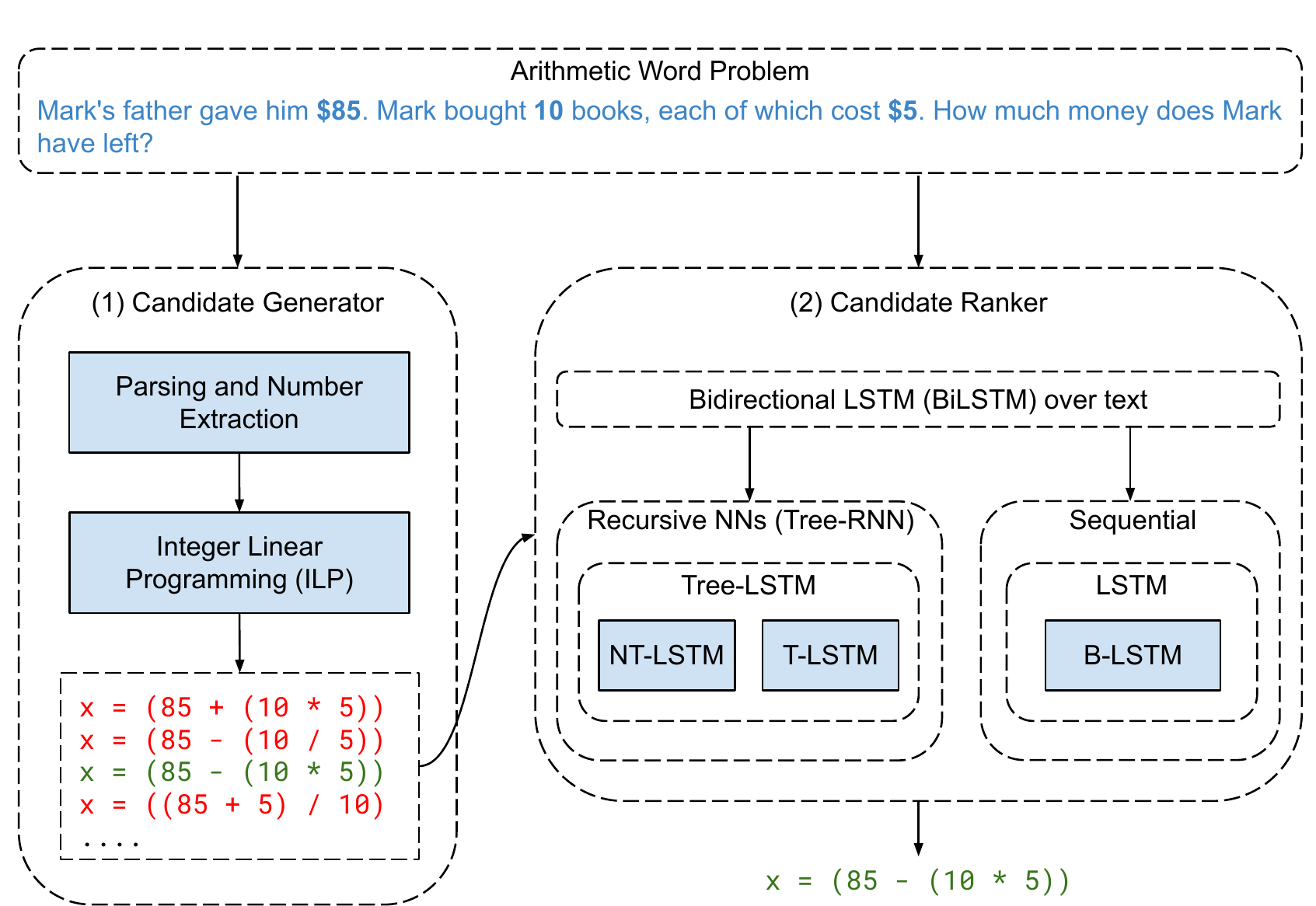}
\captionsetup{singlelinecheck=off}
\caption[test]{
High-level conceptual view of the arithmetic word problem architecture used throughout the paper. It consists of two main components:
\protect\begin{enumerate*}[(1)]
    \item\label{fig2a} \textit{candidate generator} responsible for generating candidate equations to solve a particular \textit{arithmetic word problem}, and
    \item\label{fig2b} \textit{candidate ranker}, for selecting the best candidate from the list provided by \textit{candidate generator}, using the models \NTLSTM{}, \TLSTM{}, or \BLSTM{}. 
\protect\end{enumerate*}
}
\label{fig:conceptualview}
\end{figure}

\Figref{fig:conceptualview} provides a high-level conceptual view of the interconnection between the main components of our proposed system. The processing flow consists of two main steps. In the first step, we use the \textit{candidate generator} to generate a list of potential candidate equations for solving a particular \textit{arithmetic word problem}. To achieve this, we employ the Integer Linear Programming (ILP) constraint optimization component proposed by \mbox{\cite{koncel2015parsing}} (see \secref{sec:candidate_generator}). In the second step, the candidate equations are ranked by the \textit{candidate ranker}, and the equation with the highest score is chosen as the solution to the processed \textit{arithmetic word problem} (see \secref{sec:model}). 
In this paper, we focus on this second step by exploring the impact of structural Tree-RNN-based and sequential Long Short Term Memory-based (LSTM; \cite{hochreiter1997long}) candidate equation encoding methods.
More specifically, we define two Tree-RNN models inspired by the work of \cite{tai2015improved} on \TreeLSTM{} models:
\begin{enumerate*}[label=(\roman*)]
    \item \TLSTM{} (Child-Sum \TreeLSTM{}), and 
    \item \NTLSTM{} (N-ary \TreeLSTM{})
\end{enumerate*}. In the rest of the manuscript we refer to the general tree-structured architecture of these models as \TreeLSTM{}.
The main difference between the two is that, while in \TLSTM{} the child node representations 
are summed up, in \NTLSTM{} they are concatenated. Unlike the representation used in \cite{tai2015improved}, where the input is given by the word embeddings, our Tree-LSTM models also take as input the operation embeddings (in inner nodes) that represent each of the arithmetic operators ($-$, $+$, $\div$, $\times$). This allows our architecture to distinguish between different operators that are contained in a particular expression tree. We show that \NTLSTM{} is more suitable to deal with equations that involve non-commutative operators because this architecture is able to capture the order of the operands.
We also compare our \TreeLSTM{} models with a sequential LSTM model which we call \BLSTM{}. 
All the models (\TLSTM{}, \NTLSTM{}, and \BLSTM{}) take as input the contextualized representation of the numbers in text produced by a bidirectional \LSTM{} layer (\biLSTM{}) (see \secref{sec:model} for details).
After conducting a thorough multi-fold
experimentation phase involving multiple random weight re-initializations in order to ensure the validity of our results, we will show that the main added value of our \TreeLSTM{}-based models compared to state-of-the-art methods lays in an increased performance for 
more complex arithmetic word problems.

More concretely, 
our contribution is three-fold:
\begin{enumerate*}[(i)]
\item we propose using \TreeLSTMs{} for solving arithmetic word problems, to embed structural information of the equation,
\item we compare it against a strong neural baseline model (\BLSTM{}) that relies on sequential LSTMs,
and
\item we perform an extensive experimental study on the SingleEQ dataset, showing that our \TreeLSTM{} model achieves an overall accuracy improvement of 3\%, including an increase 
$>$15\% for more complex problems (\ie requiring multiple and non-commutative operations), compared to previous state-of-the-art results. 
\end{enumerate*}

\section{Related work}
\label{sec:related}

\noindent Over the last few years,
there has been an increasing interest in building systems to solve \textit{arithmetic word problems}. The adopted approaches 
can be grouped in three main categories: \begin{enumerate*}[label=(\roman*)]
    \item Rule-based systems,
    \item Statistical systems, and
    \item Neural network systems
\end{enumerate*}. 

\noindent\textbf{Rule-based systems}: 
The first attempts to solve
arithmetic problems date back to the 1960s 
with the work by \cite{bobrow1964natural}, who proposed and implemented 
STUDENT, a rule-based parsing system to extract numbers 
and operations between them by using pattern matching techniques. 
 \cite{charniak1968carps, charniak1969computer} extended STUDENT by including basic coreference resolution and capability to work with rate expressions (\eg ``kms per hour"). 
On the other hand, \cite{fletcher1985understanding}
designed and implemented
a system that given a propositional representation of a math problem\footnote{With propositions such as \textit{GIVE Y X P9}, where entity Y gives to entity X the object defined in P9. This proposition in particular can be linked to the first sentence of example in \figref{fig:example}: ``Mark's father gave him \$85", where Y represents ``Mark's father", X represents ``him" which is coreferenced to ``Mark", and P9 represents ``\$85" that are being given. }, applies a set of rules to calculate the final solution. 
The disadvantage of this system is that it needs a 
parsed propositional 
representation of a problem as input and cannot operate directly on raw text. 
This issue was tackled
by \cite{bakman2007robust}, who developed 
a schema-based system that consisted of six main reasoning schemas, each one with slots to fill in.  
After instantiating the schemas for a particular math problem using lexical verb-based rules, the system could derive the corresponding mathematical equation to solve the problem.  

The main disadvantages of such rule-based approaches are that they \begin{enumerate*}[label=(\roman*)]
    \item rely on hard-coded lexico-grammar rules, and 
    \item lack an integrated view of the problem to be solved, extracting operations one by one
\end{enumerate*}. We address these issues by proposing a model that integrates the mathematical representation of a problem in a single structured expression tree. This way, we are able to capture the operator-operator and number-operator relations involved in a particular mathematical expression in a unified manner. Furthermore, we 
avoid
the use of lexico-grammar hard-coded rules (\eg the use of pattern-based matching) when connecting numbers with the operators, replacing them by composition-semantic representations that link the arithmetic operations with parameters (numbers or other operations) in a recursive tree.
Consequently, our solution is more generalizable by not depending on explicit hand-crafted logic. 

\noindent\textbf{Statistical systems}: 
Recently, there has been a shift towards statistical feature-driven systems that automatically produce models by capturing patterns present in arithmetic word problem datasets.  
For example, \cite{hosseini2014learning}
 presented an 
 inductive model that links specific lexicon-based features (\eg verb categories) to equation operators. The mathematical solution to the problem is built sequentially using state transitions related to operators that are triggered by different verb categories found in the problem statement. On the other hand, \cite{mitra2016learning} connected carefully designed features to equation templates in order to solve specific problem types.  
While these techniques produced competitive results, they were limited to addition ($+$) and subtraction ($-$) operations 
on a very narrow problem set domain. 
In order to solve more diverse types of problems that also involve multiplication and division operators,
the community shifted towards more integrated approaches involving 
tree structure representations.
\cite{koncel2015parsing}  
proposed to rank candidate expression trees by training jointly a \textit{local} model to link spans of text with operator tree nodes, and a \textit{global} model that is used to score the consistency of an entire tree. The list of candidates to these two models is generated by an ILP constraint optimization component that, given a set of extracted numbers from a arithmetic word problem text as input, produces a set of candidate solution equations. 
Conversely, \cite{roy2015solving, roy2017unit} introduced the concept of \textit{monotonic expression tree} to generate candidates. It defines a set of conditions (\eg two division and subtraction nodes cannot be connected to each other) that considerably restricts the expression tree search space. The authors propose to score the resulting monotonic expression trees jointly by summing up the scores of different classifiers related to a specific expression tree (\eg 
the mathematical operator between two numbers in the tree, 
whether a particular number is related to a rate such as ``kms per hour", etc).
Recently, the same authors \citep{roy2018mapping} included additional latent declarative rules (\eg $[\mathrm{Verb1} \in \mathrm{HAVE}] \land [\mathrm{Verb2} \in \mathrm{GIVE}] \land [\mathrm{Coref(Subj1,Subj2)}] \implies \mathrm{Subtraction}$)
to link textual expression patterns (derived from preliminary
dependency parsing) to specific operations. While these statistical approaches rely on tree structures to evaluate the mathematical expressions, 
on one hand, they require high manual effort to engineer the features and, on the other hand, it is hard to scale the features to capture operations between more than two numbers. This makes it challenging to apply such models to more complex equations that involve multiple operators. 
We tackle this problem by defining a single Tree-RNN structure that evaluates an entire mathematical expression at once. This is done by recursively combining the information from the child nodes in the expression tree and then using a backpropagation mechanism to correspondingly adjust the weights of our model. Furthermore, our equation ranking architecture does not depend on hand-crafted features and parsing-dependent rules, making it more effective in generalizing across different domains. 

\noindent\textbf{Neural network systems}: 
Recently, as in all sub-domains of natural language processing, neural network architectures have been applied to tackle math word problems. The first contribution was made by 
\cite{wang2017deep}, who introduced a model trained to map problem statements to equation templates.
Their model
was expanded upon by \cite{huang2018neural}, who introduced an attention-based copy mechanism for tokens representing numbers. They used  
a reinforcement learning setting, where positive 
rewards
were assigned when the predicted mathematical expression resulted in a correct answer.
Recently, \cite{chiang2019semantically} used stack structures inside a sequential encoder-decoder setting where the encoder captures the semantics of a math word problem in a vector that is used by decoder to generate the equation to solve the problem. 
Moreover, \cite{wang2018mathdqn} proposed the
use of Q-Networks in order to generate expression trees, by giving positive reward 
whenever the 
operator between two numbers is correct. 
The aforementioned studies, while showing promising results, 
were not designed to naturally capture the structural form of mathematical expressions when multiple operators are involved (\eg $1+(2/3)$ \vs $(1+2)/3$). 
We propose encoding equations with \TreeLSTMs{}~\citep{tai2015improved}, which are recursive neural sequence models, thus allowing to naturally reflect the execution order of operations in an expression tree by recursively combining the children nodes' semantic representations. 

\renewcommand\baselinestretch{1}{\begin{table}[t]
\centering
\resizebox{0.95\textwidth}{!}{
\begin{tabular}{lccccc}
\toprule
\textbf{Method} & 
\textbf{Rules} & \textbf{Features} & \textbf{N-Nets} & \textbf{Tree-Based} & \textbf{Tree-Based}   \\
 & & & & \textbf{Representation} & \textbf{Encoding}  \\
 \midrule
 \cite{bobrow1964natural} & \yesmarker & \nomarker & \nomarker & \nomarker & \nomarker \\
 \cite{charniak1968carps, charniak1969computer} & \yesmarker & \nomarker & \nomarker & \nomarker & \nomarker \\
 \cite{fletcher1985understanding} & \yesmarker & \nomarker & \nomarker & \nomarker & \nomarker \\
 \cite{bakman2007robust} & \yesmarker & \nomarker & \nomarker & \nomarker & \nomarker \\
 \cite{hosseini2014learning} & \nomarker & \yesmarker & \nomarker & \nomarker & \nomarker \\
  \cite{koncel2015parsing} & \yesmarker & \yesmarker & \nomarker & \yesmarker & \nomarker \\ 
 \cite{mitra2016learning} & \nomarker & \yesmarker & \nomarker & \nomarker & \nomarker \\
 \cite{roy2015solving, roy2017unit} & \nomarker & \yesmarker & \nomarker & \yesmarker & \nomarker \\ 
 \cite{wang2017deep} & \nomarker & \nomarker & \yesmarker & \nomarker & \nomarker \\    
 \cite{roy2018mapping} & \yesmarker & \yesmarker & \nomarker & \yesmarker & \nomarker \\ 
 \cite{huang2018neural} & \nomarker & \yesmarker & \yesmarker & \nomarker & \nomarker \\ 
 \cite{wang2018mathdqn} & \nomarker & \yesmarker & \yesmarker & \yesmarker & \nomarker \\
 \cite{chiang2019semantically} & \nomarker & \nomarker & \yesmarker & \nomarker & \nomarker \\
 \cite{li2019modeling} & \nomarker & \nomarker & \yesmarker & \nomarker & \nomarker \\
 Our Approach (\TLSTM{} \& \NTLSTM{}) & \nomarker & \nomarker & \yesmarker & \yesmarker & \yesmarker \\   
\midrule
\end{tabular}
}
    \caption[test]{Comparison of the various architectures explored in related work. We focus on the following five characteristics: \protect\begin{enumerate*}[(i)] 
	 \item \textit{Rules} indicates whether a rule-based approach is used or not, 
	 \item \textit{Features} specifies whether the architecture relies on manually engineered features, 
	 \item \textit{N-Nets} indicates whether artificial neural networks are used or not, 
	 \item \textit{Tree-Based Representation} groups 
	 the models that incorporate information coming from tree structures (\eg by using trees for feature engineering), and
	 \item \textit{Tree-Based Encoding} indicates whether the tree structures are used as encoders in a neural network model.
	 \end{enumerate*} The {\yesmarker} indicates the presence of a particular characteristic.}
    \label{tab:comparative_related_work}
\end{table}}

\tabref{tab:comparative_related_work} compares our approach (the use of \TreeLSTM{}-based \TLSTM{} and \NTLSTM{} models) with the rest of the methods described in this Section.
The main difference of our architecture is that we explore the impact of using tree-based neural encoding (\ie by means of Tree-LSTM models). We hypothesize that this approach allows
to better capture the arithmetic equation structure than the currently predominant neural sequential models \citep{wang2017deep,wang2018translating, chiang2019semantically}. Furthermore, the independence from feature-based and rule-based methods makes our solution more generalizable. This is because our model does not depend on 
hand-crafted rules or features to capture the patterns of a particular dataset.
This aspect will be explored further when comparing the performance of our model to the current feature-based state-of-the-art system \citep{koncel2015parsing} in \secref{sec:results}. 

\textbf{Tree-RNN} models \citep{socher2011parsing} have been shown to perform better for modeling data on tasks that have an inherently hierarchical structure. For example, \cite{socher2011parsing} proposed to use recursive models in order to model the compositional structure of scene images (\eg a scene image of a house can be split in composing regions such as doors, windows, walls, etc.). The authors show that a Tree-RNN-based architecture outperforms previous methods in prediction of hierarchical structure of scene images and in scene image classification. Later, \cite{socher2013parsing} also showed how recursive structures can be used to encode the inherently hierarchical phrase structural grammar (\eg the sentence ``riding a bike" can be decomposed in the verb ``riding" and the noun phrase ``a bike", which itself can be decomposed into determiner ``a" and the noun ``bike"). This way, the authors achieved state-of-the-art performance in grammatical parsing of the sentences. More recently, \citet{tai2015improved} and \citet{chen2017enhanced} showed how encoding the syntactic parsing trees of the sentence with 
Tree-LSTM
models can improve the performance in tasks such as sentiment classification and semantic relatedness (\eg natural language inference). Similarly, we propose to take advantage of the inherently hierarchical representation of mathematical expression trees by encoding them using 
Tree-LSTM
architectures. Our experiments demonstrate that this representation can be helpful in capturing the semantic relations between operators needed in order to solve more complex arithmetic problems consisting of multiple and/or non-commutative operations.

\section{Proposed Architecture}
\label{sec:architecture}
\noindent 
Shortly stated, our task at hand is to identify the correct arithmetic equation, corresponding to an arithmetic problem expressed in natural language text.
We follow a two-step approach 
similar to the work of
\cite{koncel2015parsing}, which formalizes solving multi-sentence arithmetic word problems as 
\begin{enumerate*}[(i)]
\item the generation and 
\item ranking
\end{enumerate*}
of expression trees.
The {first step} consists of generating candidate equations using the ILP optimization solver proposed in \cite{koncel2015parsing} (\textit{candidate generator} component in \figref{fig:conceptualview}). 
The {second step} ranks these candidates and selects the top ranked one as the final answer to the arithmetic word problem (\textit{candidate ranker} component in \figref{fig:conceptualview}). We use the rest of this section to provide more insights into the \textit{candidate generator} component in~\secref{sec:candidate_generator}, and to describe in detail our proposed \textit{candidate ranker} model in~\secref{sec:model}.

\subsection{Candidate Generator}
\label{sec:candidate_generator}

\noindent This component is responsible for
generating possible candidate equations 
to solve a given arithmetic word problem. 
A straightforward solution would be to perform an exhaustive search on all the possible arithmetic expression trees given $n$ extracted numbers from a particular problem.
 However, the resulting search space would grow exponentially with $n$, which makes this approach not scalable.
In order to deal with this exponential growth in the number of candidates, we re-use the Integer Linear Programming (ILP) solver proposed by \cite{koncel2015parsing}. This solver takes as input the extracted numeric quantities with extra attributes derived from syntactic parsing\footnote{Stanford Dependency Parser in CoreNLP 3.4 is used.}, and generates the most promising candidate equations using two types of constraints:  
\begin{enumerate}
    \item \textit{Hard Constraints:} such as the maximum equation length and syntactic validity of equations (\eg only one unknown allowed, no division by 0, etc.). As a post-processing step, the ILP solver also removes the arithmetic expressions that produce negative or fractional results.
    \item \textit{Soft Constraints:} these constraints assign additional weight to candidate equations whose related entity types (extracted from dependency parse tree) are consistent. For example, in the problem of \figref{fig:example}, the sum (85 + 5) will be prioritized over the sum (5 + 10), because both 85 and 5 refer to the same entity type (``\$''), while 10 refers to entity type ``books''.
\end{enumerate}
\noindent To provide a fair comparison between the \textit{candidate ranker} model of ALGES proposed by \cite{koncel2015parsing} and our approach (see \secref{sec:model}), we use both
the same constraint configuration, and
also consider only the top 100 equations produced by the candidate generator.
As in ALGES, we report the coverage as \textit{\MaxILP{}} in our results section (see \secref{sec:results}). 
Additionally, we include in our result tables the performance of the \textit{\TopILP{}} approach, which consists of selecting the highest scored candidate by the ILP solver. 
This score allows us to estimate the impact of the \textit{candidate ranker} component.

\subsection{Candidate Ranker}
\label{sec:model}
\begin{figure}[!ht]
\centering
\includegraphics[width=.60\columnwidth,clip]{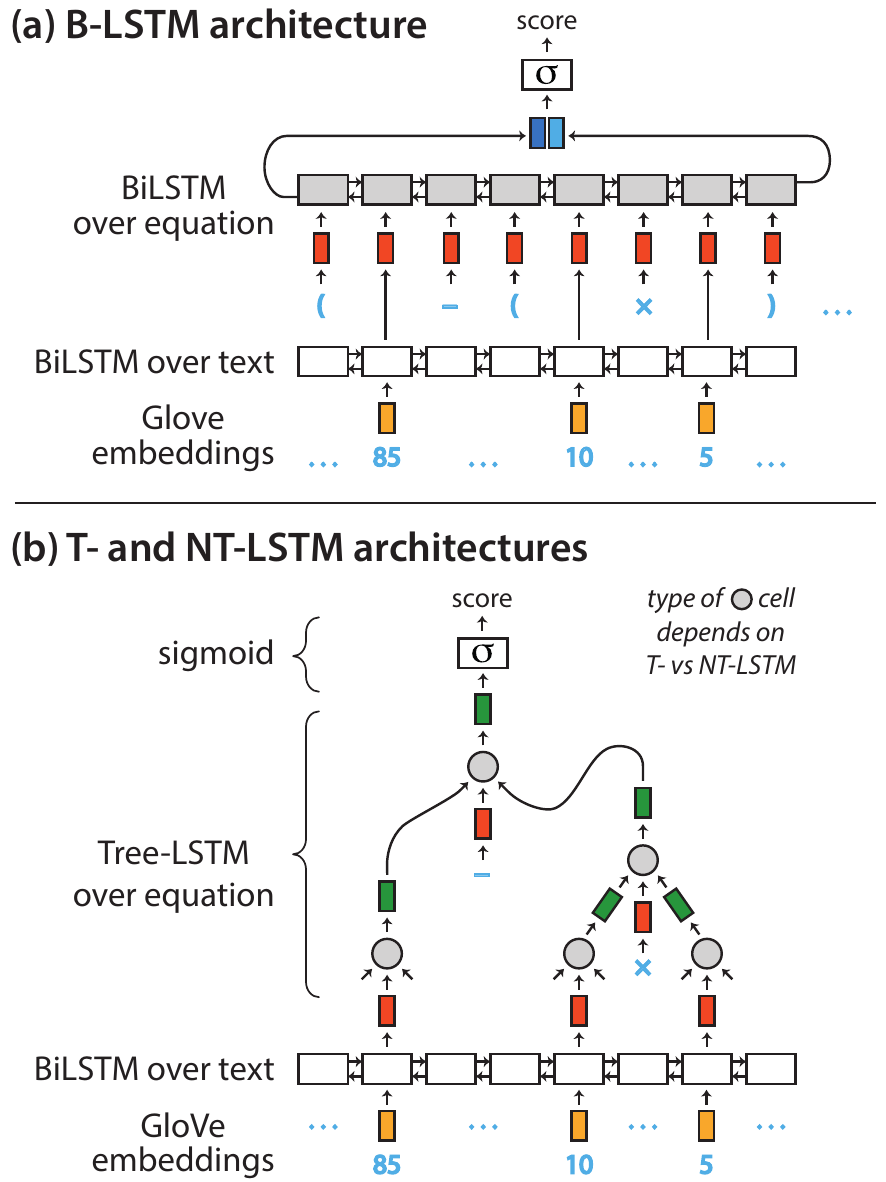}
\captionsetup{singlelinecheck=off}
\caption[test]{Models for scoring equations, taking the text and the equation from \figref{fig:example} to score (\eg $85 - (10 \times 5)$) as input:
    \protect\begin{enumerate*}[(i)]
    \item a word embedding layer at the bottom,
    \item a \biLSTM{} layer over the text, and
    \item a top layer that encodes the equation.
    \protect\end{enumerate*}
 For the latter we consider either
  \protect\begin{enumerate*}[label=\textbf{(\alph*)}]
     \item \label{arch:bilstm} a sequential \biLSTM{} (\BLSTM{} architecture), 
     or
     \item \label{arch:treelstm} a structured \TreeLSTM{} (\TLSTM{} and \NTLSTM{} architectures).
  \protect\end{enumerate*}
}
\label{fig:model}
\end{figure}

\noindent Our proposed candidate ranker model architecture is sketched in~\figref{fig:model} and comprises: 
\begin{enumerate*}[(i)]
    \item a word embedding layer,
    \item a bidirectional \LSTM{} layer (\biLSTM{}) over the text, and
    \item an additional layer that encodes the equation, using either \biLSTM{} (\BLSTM{} model) or \TreeLSTM{} (\TLSTM{} and \NTLSTM{} models) based approaches, detailed below.
\end{enumerate*}

The input to our model is a \textcolor{wordcolor}{\textbf{sequence of tokens}} of length $N$, $W=\{w_1,..., w_N\}$ of the arithmetic word problem, which we pass through an \textcolor{embeddingcolor}{\textbf{embedding layer}} to obtain embedded representations $X=\{x_1,..., x_N\}$ where 
$x_t \in \mathbb{R}^{d_1}$.
We adopt a \biLSTM{} to obtain \textcolor{biLSTMcolor}{\textbf{contextual representations}} of the tokens. The following is the formal representation of the first \LSTM{} \citep{hochreiter1997long} layer used to produce the representation referred to as ``\textit{\biLSTM{} over text}'' in \figref{fig:model}: 

\begin{align}
    i_t = \sigma \left(W_i x_t + U_i h_{t-1} + b_i \right) \label{lstm:1} \\ 
    o_t = \sigma \left(W_o x_t + U_o h_{t-1} + b_o \right) \label{lstm:2} \\ 
    f_t = \sigma \left(W_f x_t + U_f h_{t-1} + b_f \right) \label{lstm:3} \\ 
    u_t = \mathrm{tanh} \left(W_u x_t + U_u h_{t-1} + b_u \right) \label{lstm:4} \\
    c_t = f_t \odot c_{t-1} + i_t \odot u_t  \label{lstm:5}
    \\
    h_t = o_t \odot \mathrm{tanh}(c_t)  \label{lstm:6}
\end{align}
\noindent where $t \in \{1,..., N\}$ represents a particular recursive execution time step and
$h_t \in \mathbb{R}^{d_2}$ is the \LSTM{} hidden state. The advantage of using the \LSTM{}-based structure instead of a simpler recursive formulation, such as $h_t = \tanh (W x_t + Uh_{t-1} + b)$, is that an \LSTM{} model avoids the problems of exploding or vanishing gradients during the training process discussed in \cite{hochreiter1997long,bengio1994learning}. This is achieved by using additional weight matrices 
and \textit{gates} $\sigma$ in \equsrefrange{lstm:1}{lstm:3} in order to regulate the amount of information from previous execution steps $h_{t-1}$ and current input $x_t$ that affect the current state $h_t$.\footnote{For a more detailed description of the \LSTM{} architecture please refer to \cite{hochreiter1997long}.}
More concretely, $W_i$, $W_f$, $W_o$, $W_c \in \mathbb{R}^{d_2\times d_1}$ and $U_i$, $U_f$, $U_o$, $U_c \in \mathbb{R}^{d_2\times d_2}$ are the weight matrices related to different LSTM gates, and $b_i$, $b_f$, $b_o$, $b_c \in \mathbb{R}^{d_2}$ are the respective biases. 
In our experiments we initialize $x_t$ with GloVe word embeddings \citep{pennington:14} and keep them static during training. These \textit{GloVe embeddings} 
are depicted at the bottom of graphs \textbf{(a)} and \textbf{(b)} in \figref{fig:model}.
In order to obtain the \biLSTM{} representation (\textit{``\biLSTM{} over text''} in \figref{fig:model}), we run two \LSTMs{} in different directions and concatenate the respective hidden states. This results in $N$ hidden state representations $H=\{h_1^{(b)}, ..., h_N^{(b)}\}$ where $h_i^{(b)} \in \mathbb{R}^{d_3}$ and $d_3 = 2 \cdot d_2$. Using the input in $H$, we propose two different models to encode the candidate equations referred to as \textbf{(a)} and \textbf{(b)} in \figref{fig:model}, and explained below: 

\noindent \textbf{\ref{arch:bilstm} Sequential \BLSTM{}}: We perform an in-order traversal of the expression tree to obtain a sequential representation of the equation (\eg $\left(85-(10\times5)\right))$
that is encoded using a second \biLSTM{} (see \textit{``BiLSTM over equation''} in \figref{fig:model}\ref{arch:bilstm}). 
We use as input the hidden state representations $H$ calculated above for the numbers and (trainable) embeddings 
$O \ =\{o_{-}, o_{+}, o_{\div}, o_{\times}, o_{(}, o_{)} \}$ for the operators ($-, +, \div, \times$) and opening/closing parentheses. More formally, the input to \biLSTM{} is represented by 
$X^E = \{x^e_1, ..., x^e_K\}$ where $x^e_t \in \{ H \cup O \}$ , $x^e_t \in \mathbb{R}^{d_3}$
and $K$ is the number of tokens in the equation, including parentheses and operations. E.g., the equation $(85-(10 \times 5))$ contains $9$ tokens. In terms of the formal notation of LSTM in \equsrefrange{lstm:1}{lstm:6}, each $x^e_t$ corresponds to input vector $x_t$. In order to obtain a score for ranking the equation, we concatenate the last (left and right) hidden states of the \biLSTM{} producing a vector of dimensionality $d_4$, and then apply a linear transformation followed by a \textit{sigmoid} function.  

\noindent \textbf{\ref{arch:treelstm} \TreeLSTM{}}: 
We base our implementation on the \TreeLSTM{} architecture proposed by \cite{tai2015improved}.
This architecture is based on the LSTM formulation described in \equsrefrange{lstm:1}{lstm:6}, but instead of being linearly linked, the input to a particular \LSTM{} cell can come from different child step \LSTM{} executions. More formally, we can describe the \TLSTM{} structure as follows: 

\begin{align}
    \tilde{h}_t &= \sum_{k\in \{L, R\}}{h_{t-1}^{k}} \label{tlstm:0} \\
    i_t &= \sigma \left(W_i x_t + U_i \tilde{h}_t + b_i \right) \label{tlstm:1} \\ 
    o_t &= \sigma \left(W_o x_t + U_o \tilde{h}_t + b_o \right) \label{tlstm:2} \\ 
    f_{t}^{k} &= \sigma \left(W_f x_t + U_f h_{t-1}^k + b_f \right) \label{tlstm:3} \\ 
    u_t &= \mathrm{tanh} \left(W_u x_t + U_u \tilde{h}_t + b_u \right) \label{tlstm:4} \\
    c_t &= i_t \odot u_t + \sum_{k\in \{L,R\}}{f_{t}^k} \odot c_{t-1}^k \label{tlstm:5} \\
    h_t &= o_t \odot \mathrm{tanh}\left(c_t\right) \label{tlstm:6}
\end{align}

\noindent where $\{L,R\}$ is the set that consists of left ($L$) and right ($R$) child nodes for the current execution node at step $t$. More specifically, a particular execution step $t$ corresponds to the respective arithmetic operation in the expression tree (see \figref{fig:example}\ref{fig1c}). This step takes as input the cell ($c$) and hidden ($h$) states of previous execution step ($t-1$) for each of the child nodes ($\{L,R\}$) that correspond to left and right operands in the expression tree. This execution process is recursive: each of the execution steps produces as output a hidden state $h_t$ (\equref{tlstm:6}) which is used by the parent execution step recursively in \equref{tlstm:0} either as left ($h_{t-1}^L$) or right ($h_{t-1}^R$) child. Additionally, a \textit{cell state} $c_t$ is passed across the execution steps, and contains a summarized historic information of the tree traversal\footnote{Post-order traversal is used, since it reflects the order of operator execution in an arithmetic equation to obtain the final result.} operations performed so far. Similarly as with LSTM, a \textit{forget} gate $f_{t}^k$, \textit{input} ($i_t$) and \textit{update} ($u_t$) gates are used to determine which historic information is kept (forget gate) and which new information is added (input/update gates) to the cell state.
 $W_i$, $W_o$, $W_f$, $W_u \in \mathbb{R}^{d_4 \times d_3}$ together with $U_i$, $U_o$, $U_f$, $U_u \in \mathbb{R}^{d_4 \times d_4}$ are the weight matrices that transform the inputs $x_t \in \mathbb{R}^{d_3}$, the current hidden state $\tilde{h}_t \in \mathbb{R}^{d_4}$ and the children's hidden states 
 $h_{t-1}^k \in \mathbb{R}^{d_4}$, by means of
 the \TreeLSTM{} gate representations. As depicted in \figref{fig:model}\ref{arch:treelstm}, the inputs $x_t$ 
 to the leaf nodes are the hidden state representations in $H$ (coming from \textit{``BiLSTM over text''} in \figref{fig:model}\ref{arch:treelstm}) on the positions where the numbers occur in the problem statement. The input $x_t$ to the inner nodes, on the other hand, are one of the randomly initialized operation embeddings $O = \{o_{-}, o_{+}, o_{\div}, o_{\times} \}$ 
 depending on the operation represented by the node. 
 This contrasts with the original setup proposed in \cite{tai2015improved} where the input $x_t$ always comes from the word representation in the sentence. By using a separate operation embeddings set $O$ as input, we expect our model to be able to capture a semantic representation for each of the different operations $o \in O$.
 The \TreeLSTM{} model finally outputs the hidden state for the root of the expression tree (\ie the last executed operation), which is then passed through a sigmoid to deliver the score for a particular candidate arithmetic expression.

While \TLSTM{} allows to encode the equation information in a tree structure, it is symmetric in its child nodes. This is because 
the hidden states of the children are first summed up in \equref{tlstm:0}
 before applying the linear transformation and the gate activation functions. This could be problematic for non-commutative operations ($-$ and $\div$) where the result depends on the order of the operands. The reason for this is that \equref{tlstm:0} is commutative with respect to child nodes. Thus, given two child nodes $k \in \{L,R\}$ we have that $\tilde{h}_t = h_{t-1}^L + h_{t-1}^R = h_{t-1}^R + h_{t-1}^L$. As a consequence, the affine transformations $U_i$, $U_o$, and $U_u$ in \eqs~\ref{tlstm:1}, \ref{tlstm:2} and \ref{tlstm:4} cannot capture the order of the states of the input nodes. Furthermore, since there is only one weight matrix $U_f$ for both $h_{t-1}^L$ and $h_{t-1}^R$ in \equref{tlstm:3}, it can not apply a different affine transformation for left and right child nodes. This makes the \TLSTM{} model indifferent to the order of the arguments of the operations in a particular expression tree. 
 Therefore, we introduce a second model, called \NTLSTM{}, that uses distinct weight matrices to transform each of the children's hidden states. More formally, the gate definition in \NTLSTM{} is as follows: 

\begin{align}
    i_t &= \sigma \left(W_i x_t + \sum_{k\in\{L,R\}}{U_i^{k} h_{t-1}^{k}} + b_i \right) \label{ntlstm:1} \\ 
    o_t &= \sigma \left(W_o x_t + \sum_{k\in \{L,R\}}{U_o^{k} h_{t-1}^{k}} + b_o \right) \label{ntlstm:2} \\ 
    f_{t}^k &= \sigma \left(W_f x_t + \sum_{l \in \{L,R\}}{U_f^{kl} h_{t-1}^{l}} + b_f \right) \label{ntlstm:3} \\ 
    u_t &= \mathrm{tanh} \left(W_u x_t + \sum_{k \in \{L,R\}}{U_u^{k} h_{t-1}^{k}} + b_u \right) \label{ntlstm:4} \\
    c_t &= i_t \odot u_t + \sum_{k \in \{L,R\}}{f_{t}^k \odot c_{t-1}^k} \label{ntlstm:5} \\
    h_t &= o_t \odot \mathrm{tanh}\left(c_t\right) \label{ntlstm:6}
\end{align} 

\noindent where, similarly as for \TLSTM{}, $\{L,R\}$ is the set of child nodes. By introducing different weights $U$ for each of the child node states $h_{t-1}^{k}$, we make sure that the model can differentiate between the order of the operands. This is because now each of the affine transformations $U_i^{(l)}$, $U_o^{(l)}$ 
and $U_t^{(l)}$ is different for each input child hidden state $h_{t-1}^{l}$ in \eqs~\ref{ntlstm:1}, \ref{ntlstm:2} and \ref{ntlstm:4}. Similarly, each of the children's ($k \in \{L,R\}$) forget gates $f_t^k$ contains now two affine transformations $U_f^{kl}$ ($l \in \{L,R\}$), one for each child. This way, the model can prioritize (components of $f_t^k$ close to $1$) or inhibit (components of $f_t^k$ close to $0$) separately the input of a particular child $k$ based on the state of another child $l$ ($k \neq l$). This can be useful when the state of one of the operands (\eg influenced by the words that surround a particular number in text) has a strong indication of some operation, while the state of the other has very little evidence.
As we will show in~\secref{sec:results}, the use of \NTLSTM{} makes a big difference compared to the performance of \TLSTM{} for equations involving non-commutative operations. 

\section{Experimental setup}
\label{sec:experimental_setup}
\noindent We evaluate the proposed models 
(code publicly available\footnote{\url{https://github.com/klimzaporojets/arithmetic-word-problems}})
on the SingleEQ dataset introduced by~\cite{koncel2015parsing}. SingleEQ consists of 1,117 sentences and 15,292 words, and includes 508 arithmetic problems of varying complexity (\ie equations with single or multiple operators). Each of the word problems is mapped to a single correct equation with one unknown. These equations include one or more of the following operators: multiplication ($\times$), division ($\div$), subtraction ($-$), and addition ($+$). The data was gathered from the following grade-school websites: \url{http://math-aids.com}, \url{http://k5learning.com}, and \url{http://ixl.com} as well as from a subset of problems from \cite{kushman2014learning}.
To obtain results comparable to previous work, we perform 5-fold cross-validation using the original splits defined in~\cite{koncel2015parsing}. Similar to the work of \citet{koncel2015parsing} and \citet{wang2018mathdqn}, we report performance using the overall accuracy metric. 
The training/testing process is run for 5 different splits, in each one a separate fold is left as test set. This way, our results are reported on the whole SingleEQ dataset by concatenating the predictions of \textit{test} folds across the splits. In total, we train 25 models with different seeds (5 for each split) and report average and standard deviation in Tables~\ref{tab:results_full}--\ref{tab:results_complex} and \ref{tab:results_asym} in \secref{sec:results}. Furthermore, we tune the neural net hyperparameters independently for each of the splits on the validation set that consists of  20\% randomly selected arithmetic problems in each of the train folds. Due to limited resources that prevented us to perform a complete grid search, we conduct the hyperparameter tuning in steps. More specifically,
in each step we perform a grid search on two hyperparameters that we identified as most correlated with each other. \Tabref{tab:hyper_search_space} summarizes our hyperparameter search space for each of the sequential tuning steps. Besides the usual hyperparameters (\ie learning rate, batch size and dropout) tuning, we also adjust the dimensionalities $d_3$ (Dim LSTM) of the first BiLSTM layer (indicated as ``\textit{\biLSTM{} over text}'' in \figref{fig:model}), and $d_4$ (Dim Encoder) of either the sequential \biLSTM{}  (``\textit{\biLSTM{} over equation}'' in \figref{fig:model}) or the tree-based \NTLSTM{} models' encoder layers (``\textit{\TreeLSTM{}}'' in \figref{fig:model}). The best hyperparameters are chosen after training for 75 epochs for each of the cross-validation splits independently.
\begin{table*}[!ht]
\centering
\resizebox{0.9\textwidth}{!}{%
\begin{tabular}{c@{\hspace{.7cm}}ccccc} 
 \toprule
 \multirow{2}{*}{Step} & 
 \multicolumn{5}{c}{Hyperparameters}  \\
\cline{2-6}
  & \multicolumn{1}{c}{Learning Rate} & \multicolumn{1}{c}{Batch Size} & Dim LSTM & Dim Encoder & Dropout \\
 \midrule
1 & \{\num{3e-4}, \num{1e-4}\} & \{\num{64}, \num{128}\} & - & - & - \\
2 & - & - & \{\num{256}, \num{512}\} & - &  \{\num{0.3}, \num{0.4}\} \\
3 & - & - & - & \{\num{256}, \num{512}\} & \{\num{0.3}, \num{0.4}\} \\
\bottomrule
\end{tabular}
}
\caption{The range of the hyperparameter search space for each of the hyperparameter tuning steps for each of the cross-validation splits of SingleEQ dataset.}
\label{tab:hyper_search_space}
\end{table*}

Furthermore, we partition the dataset into several subsets to investigate the effect of varying problem complexity on the models' performances.
These different subsets are characterized in \tabref{tab:subsets}. We form three main categories:
\begin{enumerate*}[(i)]
\item \textbf{Full:} the whole dataset is included in this setting,
    \item \textbf{Complexity:} two subsets (\ie Single, Multi) are formed based on the number of operators in the solution's equation, and
    \item \textbf{Symmetry:} four main subsets, namely Single$_{\textrm{sym}}$, Single$_{\textrm{asym}}$, Multi$_{\textrm{sym}}$, and Multi$_{\textrm{asym}}$ are formed to indicate whether the solution's equation contains single/multiple symmetric ($\times$ and $+$) or asymmetric ($\div$ and $-$) operations.
\end{enumerate*}

\begin{table}[t]
    \centering
    \resizebox{0.7\columnwidth}{!}{
    \begin{tabular}{lcc}
    \toprule
    {\textbf{Subset}} & \textbf{Equation types} & \textbf{\# Problems} \\
    \midrule
    Full & All operators & 508 \\ 
    \midrule
    Single & Single operator & 390 \\ 
    Multi & Multiple operators & 118 \\ 
    \midrule
    Single$_{\text{sym}}$ & Single symmetric operators & 208 \\ 
    Multi$_{\text{sym}}$ & Multiple symmetric operators  & 68 \\ 
    Single$_{\text{asym}}$ & Single asymmetric operators & 182 \\ 
    Multi$_{\text{asym}}$ & Multiple asymmetric operators  & 50 \\ 
   \bottomrule
    \end{tabular}
    }
    \caption{The defined subsets of the SingleEQ dataset with varying degrees of complexity.}
    \label{tab:subsets}
\end{table}

We hypothesize that our \TreeLSTM{} models will exhibit stronger performance on subsets involving multiple and/or non-commutative operations (Multi, Multi$_\text{sym}$, Multi$_\text{asym}$), since they should be able to better capture the semantic relationships between operator nodes encoded in a tree structure. We also expect a significant difference between \TLSTM{} and \NTLSTM{} architectures on subsets involving non-commutative operations (Single$_\textrm{asym}$ and Multi$_\textrm{asym}$).
By using different weight matrices to transform each of the children's states (see Eqs.~\ref{ntlstm:1}--\ref{ntlstm:4} of the \NTLSTM{} in \secref{sec:model} for more details), the \NTLSTM{} model should be able to capture the order of the operands and link 
the resulting structural information of a particular
non-commutative
mathematical expression
to the semantic representation of the problem statement. 

We obtain the top-$100$ equation-trees using the ILP solver of~\cite{koncel2015parsing}, which we rank using scores provided by our proposed model (see~\secref{sec:model}). Training of our model is performed using the Adam optimizer~\citep{kingma:14}. As a bottom token representation layer, we use pre-trained 100-dimensional ($d_1 = 100$) GloVe embeddings~\citep{pennington:14}\footnote{ \url{https://nlp.stanford.edu/projects/glove/}} which we keep static during the training process. 
\section{Results}
\label{sec:results}

\noindent In this section, we evaluate the performance of our proposed models on the SingleEQ dataset. Besides the performance on the full dataset, we are particularly interested in evaluating how each architecture behaves when evaluated on arithmetic problems of varying complexity. We assume that the problems become more complex
\begin{enumerate*}[(i)]
    \item as the number of needed mathematical operators grows, and
    \item when the used operators are non-commutative (asymmetric).
\end{enumerate*}
We hypothesize that our structured \TreeLSTM{}-based approach is better suited to solve 
the aforementioned
complex problems. 
In order to demonstrate this, we perform an extensive evaluation (Tables~\ref{tab:results_full}--\ref{tab:results_complex} and \ref{tab:results_asym}) of our models on subsets of different degree of complexity as defined in \tabref{tab:subsets}. Furthermore, in all of the result tables we include the potential maximum accuracy that can be achieved when using the candidates from the ILP \textit{candidate generator} (\MaxILP{}). This allows us to estimate how much  improvement can still be achieved by \textit{candidate ranker}. Conversely, in order to evaluate the impact of \textit{candidate ranker} models, we also report the accuracy achieved when picking the top-weighted candidate by ILP solver (\TopILP{}).

\noindent\textbf{Comparison on the Full dataset}:~\tabref{tab:results_full} shows the results of the evaluated systems on the Full 
SingleEQ dataset. The proposed models are the
\begin{enumerate*}[(i)]
    \item \BLSTM{},
    \item \TLSTM{}, and  
    \item \NTLSTM{} as presented in~\secref{sec:model}. 
\end{enumerate*}
 Clearly, all newly proposed architectures outperform previous methods. 
 \begin{table}[t]
\centering
\resizebox{0.7\columnwidth}{!}{%
\begin{tabular}{@{\extracolsep{4pt}}ccccccccc@{}} 
 \toprule
 \multicolumn{1}{c}{Model} & \multicolumn{1}{c}{Features} & \multicolumn{1}{c}{Trees}  & \multicolumn{1}{c}{Accuracy ($\%$)}    \\
 \midrule
\cite{hosseini2014learning} &\cmark&\xmark& 48.00   \\ 
\cite{wang2018mathdqn} &\cmark&\cmark& 52.96  \\ 
\cite{roy2015solving} &\cmark&\cmark & 66.38  \\ 
\cite{roy2017unit} &\cmark&\cmark & 72.25  \\ 
ALGES &\cmark&\cmark& 72.39 \\
\toprule
\MaxILP{} & - & -  &  91.34  \\
\TopILP{}  & - & - &  52.56  \\
\toprule
\BLSTM{}  &\xmark&\xmark  &  74.88$\pm$0.64  \\
\TLSTM{}&\xmark&\cmark & 74.88$\pm$1.06  \\ 
\NTLSTM{} &\xmark&\cmark& \textbf{75.47$\pm$0.62} \\
\bottomrule
\end{tabular}
}
\caption{Accuracy attained by the proposed and state-of-the-art methods on the \emph{Full} 
SingleEQ dataset. The \cmark{} and \xmark{} symbols indicate whether or not a model adopts hand-crafted features (`Features') or tree-structured encoding of the equations (`Trees'). The best result is typeset in \textbf{bold}. 
}
\label{tab:results_full}
\end{table}
Concretely, our methods are able to outperform strong baselines on the task, reporting an accuracy improvement of more than 3\% without relying on hand-crafted features \citep{hosseini2014learning, koncel2015parsing, roy2015solving, roy2017unit}. As detailed later on in this section (see analysis of \tabref{tab:results_complex} and \tabref{tab:results_asym}), most of this improvement with respect to the current state-of-the-art \citep{koncel2015parsing} comes from an increased performance on the more complex arithmetic word problems that involve non-commutative and multiple operations. This supports our original hypothesis that tree-based architectures are superior in representing mathematical operations between operands, specially when the mathematical expressions involve multiple operations. The hand-crafted features, used in previous works, are usually related to terms indicating specific operations and thus if they are not detected in the data, the system cannot generalize well on out-of-domain mathematical descriptions. This also applies to recent neural-based methods (see, \eg \cite{wang2018mathdqn}) where explicitly defined features are encoded in the neural structure. Furthermore, in order to ensure the validity of the differences between our proposed approaches, we carry out a bootstrap significance analysis \citep{efron1994introduction} by sampling with replacement the results of \BLSTM{}, \TLSTM{}, and \NTLSTM{} models 10,000 times. We compare the performance with respect to the \NTLSTM{} model in \tabref{tab:results_full}. 
We observe that, while our \NTLSTM{} model seems to outperform \TLSTM{} and \BLSTM{} models, this difference in performance is not significant.

\noindent \textbf{Comparison for different problem complexity}: \tabref{tab:results_complex} compares our models with ALGES \citep{koncel2015parsing} (\ie the best performing state-of-the-art model of \tabref{tab:results_full}), for subsets of different complexity levels (defined in~\tabref{tab:subsets}).~We use bootstrap significance testing to estimate the degree of certainty between the lower performing models and the best performing one in each of the subsets. We indicate significant differences with p-values below the $1\%$, $5\%$, and $10\%$ level (respectively denoted with $\ddag$, $\dag$, and $\star$) in order to identify models performing significantly different from the best performing model in each of the subsets.

\begin{table*}[t]
\centering
\resizebox{1.0\textwidth}{!}{%
\begin{tabular}{c@{\hspace{.7cm}}cc@{\hspace{.7cm}}cc@{\hspace{.7cm}}cc} 
 \toprule
 \multirow{2}{*}{Model} & 
 \multicolumn{2}{c}{Complexity}  & \multicolumn{2}{c}{Symmetric} & \multicolumn{2}{c}{Asymmetric}  \\
\cline{2-3}
\cline{4-5}
\cline{6-7}
  & \multicolumn{1}{c}{Single} & \multicolumn{1}{c}{Multi} &Single$_{\text{sym}}$ & Multi$_{\text{sym}}$  & Single$_{\text{asym}}$ & Multi$_{\text{asym}}$   \\
 \midrule
\MaxILP{}  & 93.33 & 84.75 & 94.71 & 83.82 & 91.76 & 86.00 \\
\TopILP{}  & 56.41 & 39.83 & 53.85 & 69.12 & 59.34 & 0.00 \\
\midrule
ALGES  & 77.69$^\ddag$ & 54.70$^\ddag$ & \textbf{89.90} & 72.06 & 63.74$^\ddag$ & 30.64$^\ddag$ \\
\BLSTM{}  & 79.59$\pm$0.72 & 59.32$\pm$2.34 & 80.87$\pm$0.64$^\ddag$ & 69.12$\pm$2.08$^\ddag$ & 78.13$\pm$1.36 & \textbf{46.00$\pm$4.38} \\
\TLSTM{}  & 79.59$\pm$1.24 & 59.32$\pm$1.61 & 81.35$\pm$0.98$^\ddag$ & \textbf{72.35$\pm$1.44} & 77.58$\pm$2.72 & 41.60$\pm$2.33$^\star$ \\ 
\NTLSTM{} & \textbf{80.21$\pm$0.95} & \textbf{59.83$\pm$1.75} & 81.35$\pm$1.44$^\ddag$ & 71.17$\pm$2.20 & \textbf{78.90$\pm$2.13} & 44.40$\pm$4.96 \\
\bottomrule
\end{tabular}
}
\caption{Comparison of the proposed methods with the state-of-the-art on the SingleEQ dataset in terms of accuracy. \textbf{Bold} font indicates the best results for each subset of SingleEQ (see~\tabref{tab:subsets}). The markers $\star$, $\dag$, $\ddag$ respectively indicate the achieved bootstrap significance levels $\alpha$ \textless 0.1, \textless 0.05 and \textless 0.01 with respect to the best performing model in each of the subsets.}
\label{tab:results_complex}
\end{table*}
We observe that our newly proposed models do not significantly differ among each other for solving problems involving single (Single, Single$_\mathrm{sym}$, and Single$_\mathrm{asym}$ subsets) operations. Conversely, on the problem subset requiring multiple commutative operations in their solution (Multi$_\mathrm{sym}$), our tree-based \TLSTM{} significantly outperforms  the sequential \BLSTM{} model, suggesting a potential benefit in using tree-based models to solve the problems involving multiple operations. For the subset involving multiple non-commutative operations (Multi$_\mathrm{asym}$) the \BLSTM{} and \NTLSTM{} models outperform the \TLSTM{} model, indicating a potential limitation of the latter in dealing with non-commutative operations, due to its symmetrical structure in its child nodes (a single weight matrix is used on the sum of children's states $\tilde{h}_t$ as described in \secref{sec:model}). We were surprised by an overall good performance of our sequential \BLSTM{} model, specially on Multi$_\mathrm{asym}$ subset, where it performs on par with the potentially more expressive \NTLSTM{} model. This fact also motivated us to explore the robustness of our models against additional asymmetric noise (see further analysis in the next paragraphs corresponding to the results in \tabref{tab:results_asym}). 
 
 The results in \tabref{tab:results_complex} further show that the feature-based ALGES model has competitive performance on problems requiring single and/or non-commutative operators in the solution equations. In fact, it significantly outperforms all our models on the Single$_{\textrm{sym}}$ dataset and is only marginally outperformed by our tree-based \TLSTM{} model on Multi$_{\mathrm{sym}}$.
 This suggests that the feature-based ALGES is able to explicitly capture symmetric operations by focusing on carefully engineered features. However, we observe a large drop in performance of ALGES on problems that require non-commutative (asymmetric) operations to be solved. 
 This is showcased by a difference of more than 15\% accuracy points on Single$_{\mathrm{asym}}$ and Multi$_{\mathrm{asym}}$ subsets in \tabref{tab:results_complex}. This validates our initial intuition that feature-based models fall short to capture the reasoning necessary to address
 problems that require more complex (non-commutative and multiple) operators. 

\begin{table*}[t]
\centering
\resizebox{1.0\textwidth}{!}{%
\begin{tabular}{ccccccccc} 
 \toprule
\multirow{2}{*}{Candidates} & \multirow{2}{*}{Metric} &
 \multicolumn{7}{c}{Subsets}  \\
\cline{3-9}
    &  & \multicolumn{1}{c}{Full} & \multicolumn{1}{c}{Single} & \multicolumn{1}{c}{Multi} &Single$_{\text{sym}}$ & Multi$_{\text{sym}}$  & Single$_{\text{asym}}$ & Multi$_{\text{asym}}$   \\
 \midrule
\multirow{2}{*}{ILP}  & Correct & 2.53 & 1.44 & 6.13 & 1.89 & 7.72 & 0.92 & 3.96 \\
     & Incorrect & 12 & 2.9 & 42.08 & 2.48 & 28.43 & 3.38 & 60.64 \\
    \midrule 
\multirow{2}{*}{ILP + Asym}  & Correct & 2.41 & 1.44 & 5.62 & 1.89 & 7.66 & 0.92 & 2.84 \\ 
  & Incorrect & 15.08 & 4.06 & 51.5 & 3.57 & 35.43 & 4.62 & 73.36 \\
    \midrule 
  & $\Delta$ Correct & $-$4.74\% & 0.00\% & $-$8.32\% & 0.00\% & $-$0.78\% & 0.00\% & $-$28.28\% \\ 
  & $\Delta$ Incorrect & 25.67\% & 40.00\% & 22.39\% & 43.95\% & 24.62\% & 36.69\% & 20.98\% \\
\bottomrule
\end{tabular}
}
\caption{This table illustrates the difference in average number of \textit{Correct} and \textit{Incorrect} candidate equations per problem between the original \textit{ILP} candidate generation process and the one obtained by adding noisy equations with asymmetric operators (\textit{ILP + Asym}).}
\label{tab:asymmetric_noise}
\end{table*}

  \noindent \textbf{Robustness against asymmetric noise}: The results analyzed so far are based on scoring the candidates generated by the ILP component introduced in \cite{koncel2015parsing}. However, this component already significantly reduces the number of incorrect candidates, particularly those involving asymmetric operators (\eg by removing candidate equations that produce negative or fractional results as described in \secref{sec:candidate_generator}). In order to evaluate the robustness of the proposed models, we train and evaluate them on a noisy asymmetric candidate set where we add all possible permutations to the equations involving non-commutative operators. For example, if a particular candidate equation is $x=8/2$, we would also add $x=2/8$ to the candidate set. \Tabref{tab:asymmetric_noise} shows the statistics of the noisy dataset (ILP + Asym) with respective deltas that indicate the percentage points (\%) of increase/decrease in the average number of correct/incorrect candidate equations per problem with respect to the original ILP-generated candidate set. We observe a significant increase in the number of incorrect candidates for all subsets, as well as a drop in average number of correct equations for the subsets involving asymmetric operations (Multi and Multi$_{\mathrm{asym}}$). This is because, similarly as in the original \textit{ILP} setup, we only consider the first 100 generated candidates, which in \textit{ILP + Asym} include more incorrect equations, leaving many correct ones out. This results in a lower correct/incorrect ratio that makes it more challenging for the evaluated models to find the right mathematical expression to solve a particular problem. 
\begin{table*}[t]
\centering
\resizebox{1.0\textwidth}{!}{
\begin{tabular}{cc@{\hspace{.7cm}}cc@{\hspace{.7cm}}cc@{\hspace{.7cm}}cc} 
 \toprule
  \multirow{2}{*}{Model} & \multirow{2}{*}{Full} &
 \multicolumn{2}{c}{Complexity}  & \multicolumn{2}{c}{Symmetric} & \multicolumn{2}{c}{Asymmetric}  \\
\cline{3-4}
\cline{5-6}
\cline{7-8}
  & & \multicolumn{1}{c}{Single} & \multicolumn{1}{c}{Multi} &Single$_{\text{sym}}$ & Multi$_{\text{sym}}$  & Single$_{\text{asym}}$ & Multi$_{\text{asym}}$   \\
 \midrule
\MaxILP{}  & 91.14 & 93.33 & 83.90 & 94.71 & 83.82 & 91.76 & 84.00 \\
\TopILP{} & 52.56 & 56.41 & 39.83 & 53.85 & 69.12 & 59.34 & 0.00 \\
\midrule 
ALGES  & 68.44$^\ddag$ & 75.90$^\dag$ & 43.59$^\ddag$ & \textbf{85.58} & 61.76$^\ddag$ & 64.83$^\ddag$ & 18.36$^\ddag$ \\
\BLSTM{}     & 72.99$\pm$1.14 & \textbf{78.21$\pm$0.97} & 55.76$\pm$2.10$^\ddag$ & 83.36$\pm$1.20$^\dag$ & 71.76$\pm$3.40$^\ddag$ & 72.30$\pm$2.37 & 34.00$\pm$2.19$^\star$ \\
\TLSTM{}  & 57.95$\pm$1.34$^\ddag$ & 61.69$\pm$1.49$^\ddag$ & 45.59$\pm$1.25$^\ddag$ & 80.58$\pm$2.44$^\ddag$ & 72.65$\pm$2.20$^\ddag$ & 40.11$\pm$0.92$^\ddag$ & 8.80$\pm$0.98$^\ddag$ \\ 
\NTLSTM{}  & \textbf{73.19}$\pm$\textbf{0.93} & 76.97$\pm$1.02$^\dag$ & \textbf{60.67}$\pm$\textbf{1.15} & 80.76$\pm$2.37$^\ddag$ & \textbf{76.47$\pm$0.93} & \textbf{72.63}$\pm$\textbf{1.61} & \textbf{39.20}$\pm$\textbf{2.40} \\
\bottomrule
\end{tabular}
}
\caption{Comparison of the proposed methods with the state-of-the-art model (\ie ALGES) on the SingleEQ dataset in terms of accuracy evaluated on candidate equations generated using \textit{ILP + Asym} procedure (see \tabref{tab:asymmetric_noise}). \textbf{Bold} font indicates the best results for each subset of SingleEQ (see~\tabref{tab:subsets}). The markers $\star$, $\dag$, $\ddag$ respectively indicate the achieved bootstrap significance levels $\alpha$ \textless 0.1, \textless 0.05 and \textless 0.01 with respect to the best performing model in each of the subsets.}
\label{tab:results_asym}
\end{table*}    
\Tabref{tab:results_asym} compares our models with the best performing state-of-the-art model (\ie ALGES) on candidates generated in the \textit{ILP + Asym} setting. Compared to the results presented in \tabref{tab:results_complex}, we observe a sharp decrease in performance of the ALGES model on subsets involving multiple operations (Multi, Multi$_{\mathrm{sym}}$ and Multi$_{\mathrm{asym}}$). This demonstrates once more the weakness of this feature-based model in capturing the reasoning necessary to distinguish the order of the operands involved in equations containing multiple and non-commutative operators. 
Furthermore, we observe that the sequential \BLSTM{} model is now significantly outperformed by the tree-based \NTLSTM{} on subsets involving multiple operations to be solved (Multi, Multi$_{\mathrm{sym}}$ and Multi$_{\mathrm{asym}}$). 
This again supports our initial hypothesis that tree-structured approach is better suited to capture more complex reasoning which is necessary to solve arithmetic problems. In the \textit{ILP + Asym} candidate generation setting this is even more important because of the additional noise introduced with the incorrect candidates that involve multiple and asymmetric operations. 
Conversely, for arithmetic problems involving single operations to be solved (Single, Single$_\mathrm{sym}$, and Single$_\mathrm{asym}$ subsets), the \BLSTM{} model shows a competitive performance, surpassing the tree-based \NTLSTM{} model on problems requiring single commutative operations (Single$_\mathrm{sym}$).
Additionally, we observe an important drop in performance of \TLSTM{} model which is mainly influenced by low accuracy scores on asymmetric subsets (Single$_{\mathrm{asym}}$ and Multi$_{\mathrm{asym}}$). This is in line with our initial intuition 
  that by using a single weight matrices $U_i$, $U_o$, $U_f$, $U_u$ to transform either the sum of the children's states $\tilde{h}_t$ (see \equsrefrange{tlstm:0}{tlstm:2} and \ref{tlstm:4}) or the individual children states $h_k$ (\equref{tlstm:3}), the \TLSTM{} model is unable to distinguish the order of the operands involved in asymmetric equations. This difference is less evident in \tabref{tab:results_complex} because most of the incorrect candidates involving non-commutative operations are already filtered out by the ILP component. However, in our \textit{ILP + Asym} candidate generation setup, we make sure that for each candidate involving non-commutative operation, we also include noisy candidates with all the possible asymmetric permutations. This makes it necessary not only to detect the right operation, but also to distinguish the order of the operands, where the \TLSTM{} model fails. Finally, we observe that overall (on Full dataset) our tree-based \NTLSTM{} model 
  exhibits less variance among the different bootstrap results, 
  compared to the sequential \BLSTM{} model. This indicates that \NTLSTM{} model is less susceptible to different seed initialization during the training process, making it more robust than other proposed models (\TLSTM{} and \BLSTM{}). 
\begin{table}[t]
    \centering
    \begin{tabular}{p{25mm}p{65mm}p{20mm}}    
    \toprule
    \textbf{Type} & \textbf{Problem Text} & \textbf{\NTLSTM{}} \\
    \midrule
    \multirow{1}{25mm}{Complex reasoning (57\%)} & Seth bought 20 cartons of ice cream and 2 cartons of yogurt. Each carton of ice cream cost \$6 and each carton of yogurt cost \$1. How much more did Seth spend on ice cream than on yogurt? & $20/2-1\times6$ \\ 
    \midrule
    \multirow{1}{25mm}{Parsing and counting (22\%)} 
    & Jane's dad brought home 24 marble potatoes. If Jane's mom made potato salad for lunch and served an equal amount of potatoes to Jane, herself and her husband, how many potatoes did each of them have? & n/a \\ 
    \midrule
    \multirow{2}{25mm}{World Knowledge (21\%)} & Bert runs 2 miles every day. How many miles will Bert run in 3 weeks? & $3 \times 2$ \\ 
    \cmidrule{2-3}
    & The sum of three consecutive odd numbers is 69. What is the smallest of the three numbers? & n/a \\ 
    \bottomrule
    \end{tabular} 
    \caption{Examples of problems where our \NTLSTM{} model fails. }
    \label{tab:errors}
\end{table}
\begin{table}[t]
    \small
    \centering
    \begin{tabular}{p{50mm}p{30mm}p{30mm}}
    \toprule
    \textbf{Problem Text} & \textbf{ALGES} & \textbf{\NTLSTM{}} \\
    \midrule
   Nancy bought 615 crayons that came in packs of 15.  How many packs of crayons did Nancy buy? & $615-15$ & $615/15$ \\ 
    \midrule
    Carrie's mom gave her \$91 to go shopping. She bought a sweater for \$24, a T-shirt for \$6, and a pair of shoes for \$11. How much money does Carrie have left? & $91 + 24 + 6 + 11$ & $91 - (24 + 6 + 11)$ \\
    \midrule 
    Melanie had 19 dimes in her bank. Her dad gave her 39 dimes and her mother gave her 25 dimes. How many dimes does Melanie have now ? & $19 - 39 + 25$ & $19+39+25$ \\
    \midrule
    On Saturday, Sara spent \$10.62 each on 2 tickets to a movie theater. Sara also rented a movie for \$1.59, and bought a movie for \$13.95. How much money in total did Sara spend on movies? & $10.62 + 2\times1.59 + 13.95$ & $10.62 \times 2 + 13.95 + 1.59$ \\ 
    \bottomrule
    \end{tabular}
    \caption{Examples of problems that \NTLSTM{} provides a correct solution, but current state-of-the-art ALGES~\citep{koncel2015parsing} fails. }
    \label{tab:model_correct}
\end{table}

\noindent\textbf{Error Analysis}: In order to understand our system's weaknesses, we manually analyzed the errors that it consistently makes across different training seed instances. We grouped them into three main categories represented in \tabref{tab:errors}: \textit{complex reasoning}, \textit{parsing and counting}, and \textit{world knowledge} errors. We observe that more than half (57\%) of our system's errors are due to problems requiring \textit{complex reasoning} while the numbers have been correctly extracted from the text. This reflects the results from Tables \ref{tab:results_complex} and \ref{tab:results_asym}  that show lower performance of our models on problems requiring multiple and/or non-commutative operations. As future work to alleviate this type of problems we can complement the tree-structures using additional information such as the entities inside the sentence. For instance, in the first example illustrated in \tabref{tab:errors}, if the system would know that ``ice cream" from the second sentence represents the same concept as in the first one, it would be easier to link numbers 6 and 20. A second consistent type of error is related to \textit{parsing and counting}. It mainly happens when there are several entities involved in a problem statement and the system has to count them correctly. For instance, in the second example presented in \tabref{tab:errors}, our current system is unable to produce the correct candidate mathematical expression since it can only extract the number 24 from text. Further work in improving aspects related to parsing and entity identification in the problem statement should significantly reduce this kind of mistakes. Finally, the \textit{world knowledge} related errors account for 21\% of the total mistakes. Most of these errors are due to the fact that the system is unable to capture the units correctly (\ie there are 7 days in a week, or one dime equals 0.1 dollars). However, as in the second example, some of the problems require a more advanced conceptual world understanding, such as the notion of odd numbers. Future work can be directed towards methods that are able to capture and represent this kind of world knowledge.  

\noindent\textbf{Limitations of the current state-of-the-art:} We performed an empirical study on the predicted results to understand better where our proposed model outperforms the current state-of-the art model, ALGES \citep{koncel2015parsing}. \tabref{tab:model_correct} illustrates some examples of the problems where our model gets consistently correct predictions on different training initialization weights (\secref{sec:experimental_setup}). Most of the gains came from improving on problems requiring multiple and/or asymmetric operations, corroborating our previous findings. 
\begin{table}[t]
    \small
    \centering
    \begin{tabular}{p{50mm}p{30mm}p{30mm}}
    \toprule
    \textbf{Problem Text} & \textbf{ALGES} & \textbf{\NTLSTM{}} \\
    \midrule
  Diane is a beekeeper. Last year, she harvested 2,479 \textbf{pounds} of honey. This year, she bought some new hives \textbf{and} increased her honey harvest by 6,085 \textbf{pounds}. How many pounds of honey did Diane harvest this year? & \textcolor{darkspringgreen}{$6,085 + 2,479$} & \textcolor{deepcarmine}{$6,085 - 2,479$} \\ 
  \midrule 
  Jack has a section filled with short story booklets. If each booklet has 9 \textbf{pages} and there are 49 \textbf{booklets} in the short story section, how many pages will Jack need to go through if he plans to read them all? & \textcolor{darkspringgreen}{$9 \times 49$} & \textcolor{deepcarmine}{$9 + 49$} \\ 
    \midrule
    Benny received 67 \textbf{dollars} for his birthday. He went to a sporting goods store \textbf{and} bought a baseball glove, baseball, \textbf{and} bat. He had 33 \textbf{dollars} left over. How much did he spent on the baseball gear? & \textcolor{deepcarmine}{$67 + 33$} & \textcolor{darkspringgreen}{$67 - 33$} \\
    \midrule
    Jane’s mom picked cherry tomatoes from their backyard. If she gathered 56 \textbf{cherry tomatoes} \textbf{and} is about to place them in small jars which can contain 8 \textbf{cherry tomatoes} at a time, how many jars will she need? & \textcolor{deepcarmine}{$56 + 8$} & \textcolor{darkspringgreen}{$56 / 8$} \\
    \bottomrule
    \end{tabular}
    \caption{Examples of problems that require a single operation to be solved. The first two involve commutative operations ($+$ and $\times$ respectively) where our \NTLSTM{} model fails compared to the feature-based model (ALGES; \cite{koncel2015parsing}). The rest of the examples illustrate cases where ALGES fails and \NTLSTM{} returns the correct answer. The words that represent features used in ALGES that are highly correlated with the predicted operation (\textit{entity match} and the word \textit{``and''}) are highlighted.} 
    \label{tab:model_single_eq_correct}
\end{table}

\noindent\textbf{Strengths of the current state-of-the-art and limitations of our approach:} Tables~\ref{tab:results_complex} and \ref{tab:results_asym} illustrate that in the case of single symmetric operations (Single$_{\textrm{sym}}$), the ALGES method outperforms the proposed architectures (\ie \BLSTM{}, \TLSTM{}, and \NTLSTM{}).
We hypothesize that the main reason for this is the use of carefully hand-engineered features, many of which depend on third-party tools (\eg dependency parsing).
\tabref{tab:model_single_eq_correct} illustrates four examples whose solution requires mathematical expressions with a single operator. In the first two cases our \NTLSTM{} model is outperformed by the current state-of-the-art ALGES which  correctly predicts the commutative operators ($+$ in the first example and $\times$ in the second one). We have found that these correctly predicted commutative cases are highly correlated with the \textit{entity match} feature (\ie when the noun phrase connected to the number such as ``pounds'' in the first example is the same in two numbers). This feature has high positive correlation with addition and negative correlation with multiplication operations, which is illustrated in the first and second examples respectively. It also requires an additional dependency parsing which, in case of ALGES, is performed using Stanford Dependency Parser\footnote{More concretely, the Stanford Dependency Parser in CoreNLP 3.4 is used.}. Other word-based features are also highly correlated with some operations. For example, the presence of the word ``and'' in the description of the problem is correlated with addition. However, while these features may be a strong indicators of some operators, their application is limited to problems where the underlying patterns appear. This is illustrated in the last two examples that contain two features highly correlated with the addition (\ie \textit{entity match} and ``and'' word), but that require a different (non-commutative) operation in their solutions. In both cases, biased by the most likely feature-based operation, the answer given by ALGES is incorrect. This contrasts with our feature-independent \NTLSTM{} model which manages to predict the correct equation.
This is reflected in Tables \ref{tab:results_complex} and \ref{tab:results_asym}, where the features-based approach falls short in capturing the more intricate nature of solutions involving non-commutative operations (Single$_{\mathrm{asym}}$ and Multi$_{\mathrm{asym}}$). In these cases, our tree-based \NTLSTM{} model exhibits superior performance. 

\section{Conclusion}
\label{sec:conclusion}

\noindent In this work we addressed the reasoning component involved in solving arithmetic word problems. We proposed a recursive tree architecture to encode the underlying equations for solving arithmetic word problems. More concretely, we proposed to use two different \TreeLSTM{} architectures for the task of scoring candidate equations. We performed an extensive experimental study on the SingleEQ dataset and demonstrated consistent effectiveness (\ie more than 3\% increase in accuracy on the Full dataset and more than 15\% for a subset of complex reasoning tasks) of our models compared to current state-of-the-art. 

We observed that, while very strong on simple instances involving single operations, the current feature-based state-of-the-art model exhibits a significant gap in performance for mathematical problems whose solution comprises non-commutative and/or multiple operations. 
This reveals the weakness of this method to capture the intricate nature of reasoning necessary to solve more complex arithmetic problems. Furthermore, our experiments show that, while a traditional sequential approach based on recurrent encoding implemented using \biLSTMs{} over the equation proves to be a robust baseline, it is outperformed by our recursive \TreeLSTM{} architecture to encode the candidate solution equation on more complicated problems that require multiple operations to be solved. 
This difference in performance becomes more significant as we introduce additional noise in our set of candidates by adding incorrect equations that contain non-commutative operations. 

\section*{Acknowledgment}
\noindent Part of the research leading to these results has received funding from
\begin{enumerate*}[(i)]
\item the European Union's Horizon 2020 research and innovation programme for the CPN project under grant agreement no.\ 761488, and
\item the Flemish Government under the ``Onderzoeksprogramma Artifici\"{e}le Intelligentie (AI) Vlaanderen'' programme.
\end{enumerate*}

\section*{References}
\bibliography{bibliography}
\end{document}